\PassOptionsToPackage{table}{xcolor}
\documentclass[10pt,twocolumn,letterpaper]{article}

\usepackage{cvpr}              

\usepackage{amsmath,amsfonts,bm}

\def\eqref#1{equation~\ref{#1}}

\def\1{\bm{1}}

\def\vc{{\bm{c}}}

\def\ve{{\bm{e}}}

\def\vx{{\bm{x}}}

\def\mI{{\bm{I}}}

\DeclareMathAlphabet{\mathsfit}{\encodingdefault}{\sfdefault}{m}{sl}
\SetMathAlphabet{\mathsfit}{bold}{\encodingdefault}{\sfdefault}{bx}{n}

\def\sR{{\mathbb{R}}}

\newcommand{\softmax}{\mathrm{softmax}}

\usepackage{color}
\usepackage{bm}
\usepackage{mathtools}
\usepackage{amsfonts} 
\usepackage{amsthm}
\usepackage{array,multirow}
\usepackage{tabularx}
\usepackage{arydshln}
\usepackage{adjustbox}
\usepackage{balance}
\usepackage{wrapfig}
\usepackage{pifont}
\usepackage{soul}
\usepackage{booktabs}
\usepackage{kotex}
\usepackage{makecell}
\usepackage[most]{tcolorbox}
\usepackage{listings}
\usepackage{upquote}
\usepackage{graphicx}
\usepackage{subcaption}
\usepackage[table]{xcolor}
\usepackage{colortbl}

\definecolor{cadmiumgreen}{rgb}{0.0,0.42,0.24}
\definecolor{aliceblue}{rgb}{0.94, 0.97, 1.0}
\definecolor{lightgray}{rgb}{0.95, 0.95, 0.95}
\definecolor{forestgreen}{rgb}{0.13,0.55,0.13}
\definecolor{forestred}{rgb}{0.85, 0.3, 0.3}
\definecolor{Blue9}{rgb}{0.098,0.30,0.9}
\definecolor{lightblue}{RGB}{236,244,255}

\renewcommand{\arraystretch}{1.0}
\newcolumntype{g}{>{\columncolor{lightgray}}c}

\newtcolorbox{promptbox}[1][]{%
  enhanced, breakable,
  colback=gray!3,
  colframe=gray!80,
  colbacktitle=gray!80,  
  coltitle=white,
  fonttitle=\bfseries,
  title={#1},
  top=1mm, bottom=1mm, left=1.5mm, right=1.5mm
  }

\definecolor{cvprblue}{rgb}{0.21,0.49,0.74}
\usepackage[pagebackref,breaklinks,colorlinks,allcolors=cvprblue]{hyperref}

\def\paperID{24488} 
\def\confName{CVPR}
\def\confYear{2026}

\title{Progress by Pieces: Test-Time Scaling for Autoregressive Image Generation}
\author{Joonhyung Park$^{1 *}$ \hspace{10pt} Hyeongwon Jang$^{1 *}$ \hspace{10pt} Joowon Kim$^{1}$ \hspace{10pt} Eunho Yang$^{1,2}$ \\
$^{1}$KAIST \hspace{10pt} $^{2}$AITRICS \\
{\tt\small \{deepjoon, janghw0911, kjwispro, eunhoy\}@kaist.ac.kr} \\
\normalsize{Project Homepage: \url{https://grid-ar.github.io}} 
}

\begin{document}
\maketitle

\begin{abstract}
Recent visual autoregressive (AR) models have shown promising capabilities in text-to-image generation, operating in a manner similar to large language models. While test-time computation scaling has brought remarkable success in enabling reasoning-enhanced outputs for challenging natural language tasks, its adaptation to visual AR models remains unexplored and poses unique challenges. Naively applying test-time scaling strategies such as Best-of-$N$ can be suboptimal: they consume full-length computation on erroneous generation trajectories, while the raster-scan decoding scheme lacks a blueprint of the entire canvas, limiting scaling benefits as only a few prompt-aligned candidates are generated. To address these, we introduce \textit{GridAR}, a test-time scaling framework designed to elicit the best possible results from visual AR models. \textit{GridAR} employs a grid-partitioned progressive generation scheme in which multiple partial candidates for the same position are generated within a canvas, infeasible ones are pruned early, and viable ones are fixed as anchors to guide subsequent decoding. Coupled with this, we present a layout-specified prompt reformulation strategy that inspects partial views to infer a feasible layout for satisfying the prompt. The reformulated prompt then guides subsequent image generation to mitigate the blueprint deficiency. Together, \textit{GridAR} achieves higher-quality results under limited test-time scaling: with $N{=}4$, it even outperforms Best-of-$N$ ($N{=}8$) by 14.4\% on T2I-CompBench++ while reducing cost by 25.6\%. It also generalizes to autoregressive image editing, showing comparable edit quality and a 13.9\% gain in semantic preservation on PIE-Bench over larger-$N$ baselines. The source code will be publicly released. 
\end{abstract}







\begin{figure}[t]
    \vspace{-0.02em}
    \centering
    \includegraphics[width=0.92\linewidth]{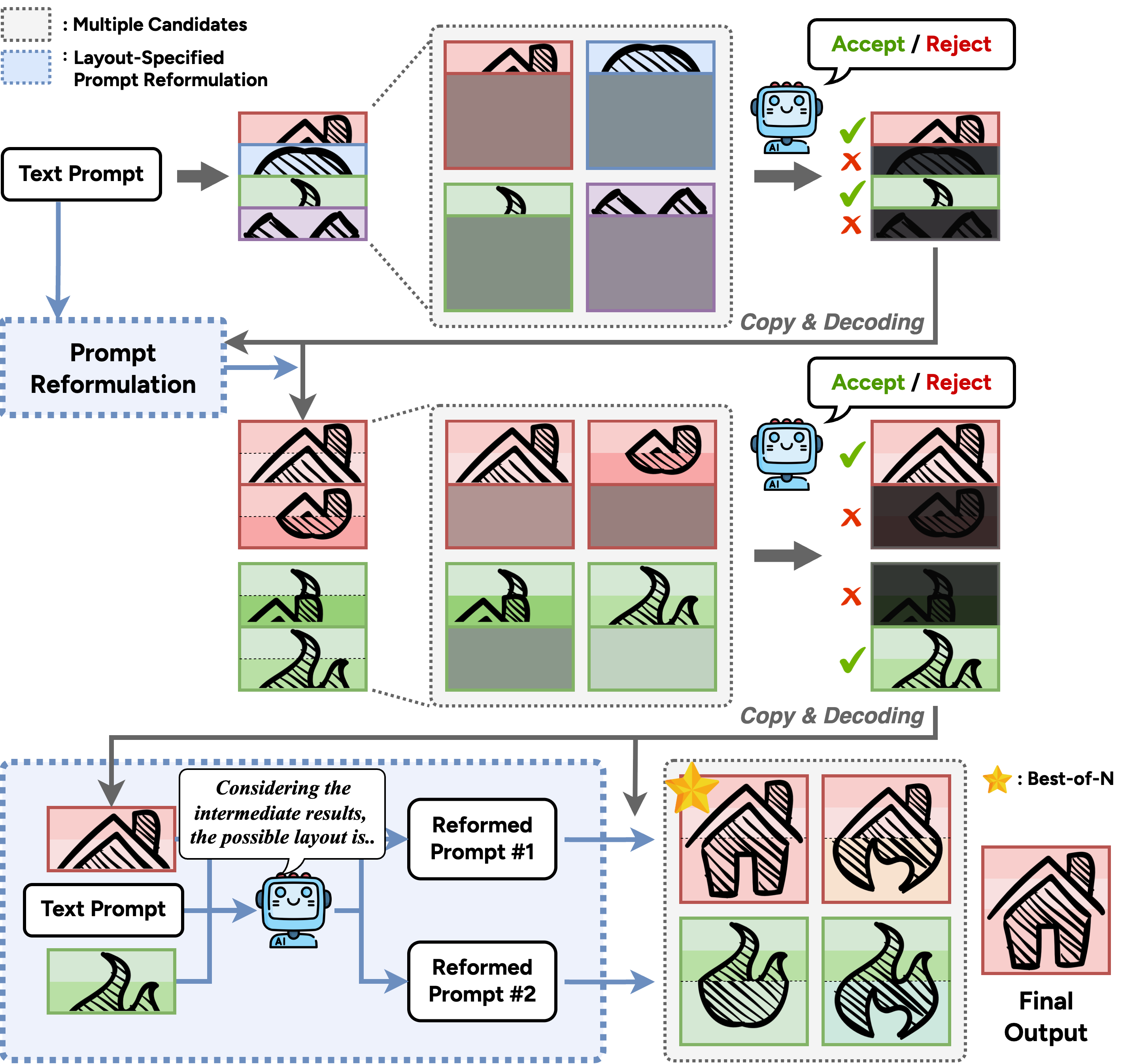}
    \vspace{-0.5em}
    \caption{A grid-partitioned progressive image generation framework (\textbf{\textit{GridAR}}) for test-time scaling of visual AR models.}
    \label{fig:thumbnail}
    \vspace{-1.5em}
\end{figure}

\vspace{-0.13in}
\section{Introduction}
Visual autoregressive (AR) models~\citep{tian2024visual,team2024chameleon,llamagen,januspro} are emerging as a compelling alternative to the long-dominant diffusion paradigm, demonstrating competitive text-to-image generation against landmark models such as DALL$\cdot$E~3~\citep{dalle3} and Stable Diffusion 3~\citep{sd3}. By encoding images as sequences of discrete tokens with the aid of VQ-VAE~\citep{van2017neural}, such models operate akin to large language models (LLMs). A growing body of research has explored raster-scan (\ie line-by-line) decoding strategies for visual AR models~\citep{wang2024emu3,llamagen,januspro,liu2024lumina}, while exploring variants such as masked modeling~\citep{xieshow,li2024autoregressive} and next-scale prediction~\citep{tian2024visual,han2025infinity}. These efforts continue to push the limits of visual fidelity in autoregressive image generation.

As LLM-style raster-scan decoding becomes feasible in text-to-image generation, a natural research question arises: how can test-time computation scaling - shown to enable human-expert level reasoning in language tasks such as math~\citep{wang2024math} and coding~\citep{chenteaching} - be applied in this setting? These strategies allocate additional computation at inference; for instance, they encourage longer, chain-of-thought (CoT) outputs~\citep{kojima2022large} or employ Best-of-\textit{N}~\citep{snell2024scaling} selection with outcome reward model (ORM) to boost the reasoning capabilities of LLMs on cognitively demanding tasks. Despite these successes in language, tailored strategies for visual AR remain underexplored, and it is still unclear how to effectively scale computation or decompose the generation process into multi-steps during test time.

In this paper, we primarily aim to devise a test-time scaling approach for visual AR models, in pursuit of achieving accurate image renderings given complex prompts, including scenarios with multiple objects, spatial relations, and attribute bindings. Prior work mostly ports LLM-style methods (reinforcement learning for token-wise CoT, CoT-augmented prompts)~\citep{t2ir1} or verifies intermediates in iterative masked AR~\citep{parm}. These approaches suggest that scaling computation at test time can help visual AR models; however, they do not fully reflect the unique characteristics that arise when images are generated by AR models. 

We highlight two key characteristics of raster-scan image generation for test-time scaling. First, due to the next-token prediction scheme, the model lacks a global blueprint of the full image. For example, when prompted with `a photo of eight bears,' if the first bear is drawn large in the upper region, the model often leaves the remaining bears undrawn in the lower region, as it also considers image fidelity. Second, the autoregressive nature makes early errors hard to fix. Consider a prompt requiring four bags: if five handles are already drawn in the upper region, the sequential generation offers no correction. These issues indicate that Best-of-$N$, a representative test-time scaling approach, is not well-suited for visual AR models: once an erroneous trajectory is initiated, it still consumes full computation, and without a global blueprint, wastes resources on misdrawn images.

Building upon this insight, we introduce the \textbf{\textit{GridAR}}, a grid-structured test-time scaling framework for autoregressive image generation. Our approach, inspired by tree-search reasoning in LLMs, focuses computation on regions where further exploration is meaningful and thereby effectively expands the search space. Specifically, the image canvas is partitioned into row-wise tiles and generates multiple candidate images for the same canvas position - \eg, four distinct upper-quarter candidates at the initial stage. 
Erroneous or infeasible candidates are then rejected, while valid ones are propagated to fill the corresponding canvas positions and serve as anchors that guide the continued generation. This \textit{glimpse-and-grow} strategy guides visual AR models to generate more sophisticated images that better follow instructions, without requiring additional training.

One natural artifact of this grid-partitioned progressive generation is a set of partial images, where only the upper portion of the canvas has been rendered. We take these as cues to address the blueprint deficiency in autoregressive models. When a vision-language model serves as a verifier to evaluate grid candidates, we simultaneously perform a \textit{layout-specified prompt reformulation}, in which the prompt is revised to explicitly encode a feasible layout grounded in the observed partial outputs. With this reformulated prompt, we propose two options: (i) apply a three-way classifier-free guidance term to steer the logits to align with the layout specified in reformulated prompts, or (2) directly substitute the prompt in subsequent generation stages, which is cost-efficient. Both approaches guide the model toward a plausible layout consistent with the intermediate results.

\begin{figure}[t]
    \vspace{-1.3em}
    \centering
    \includegraphics[width=0.83\linewidth]{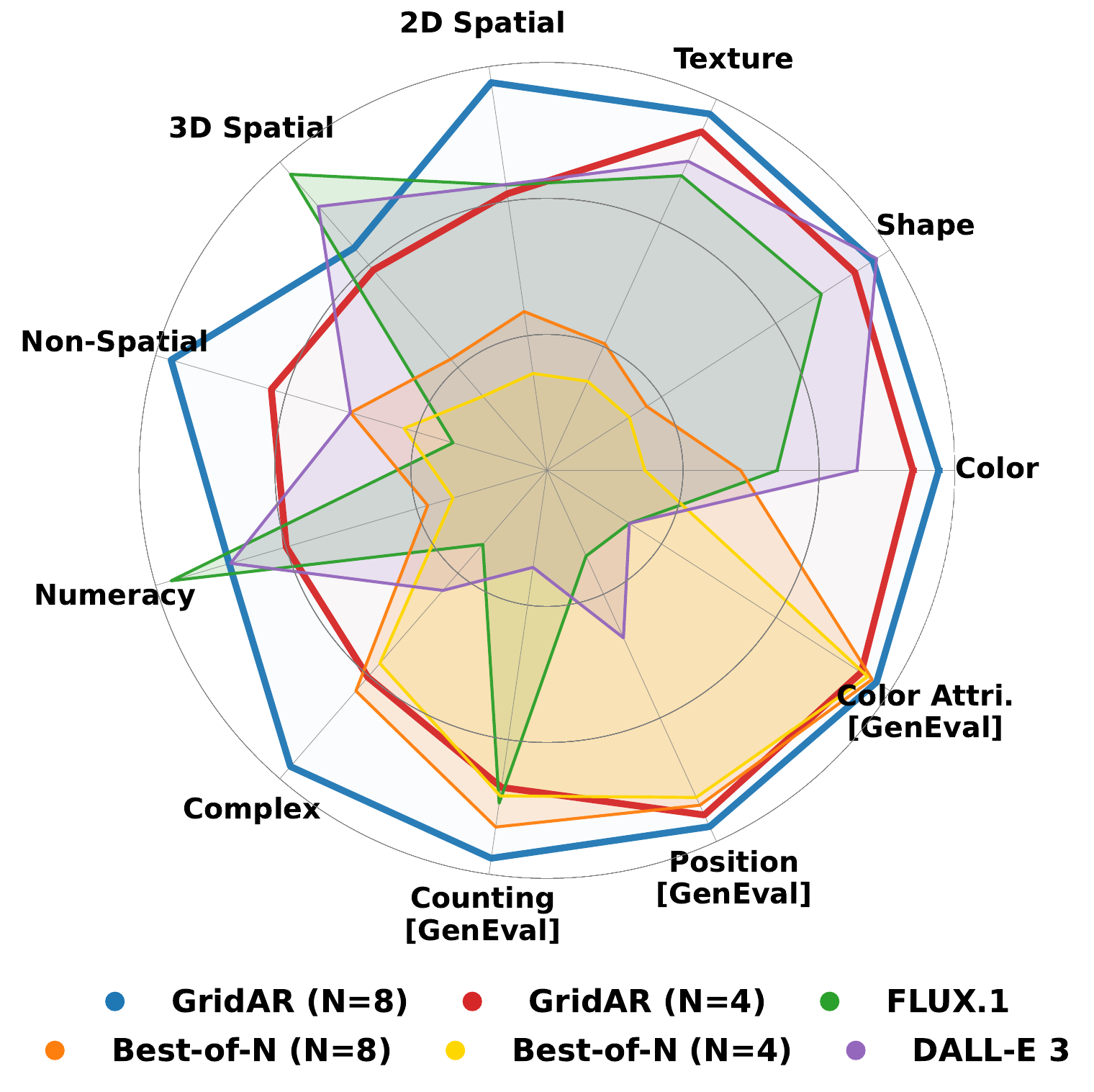}
    \vspace{-1em}
    \caption{\textbf{\textit{GridAR}} (\textit{N}=4) achieves 14.4\% higher image quality through effective test-time scaling, surpassing Best-of-$N$ (\textit{N}=8).}
    \label{fig:radar}
    \vspace{-0.2in}
\end{figure}

Our \textit{GridAR} is thoroughly validated across two tasks: text-to-image generation and image editing, using three models - Janus-Pro~\citep{janus}, LlamaGen~\citep{llamagen}, and EditAR~\citep{mu2025editAR}. Extensive experiments show that \textit{GridAR} consistently improves text-to-image generation quality from 4.8\% to 17.8\% across diverse prompt categories, and even outperforms Best-of-\textit{N} ($N$=8) using only $N$=4 candidates. In image editing, it achieves 13.9\% higher semantic preservation compared to a larger-\(N\) baseline, demonstrating a more favorable cost-performance trade-off.

In summary, our contribution is threefold:
\begin{itemize}
    \item We introduce \textit{GridAR}, a grid-structured progressive generation framework that directs computation toward promising continuations at test time, effectively expanding the search space to elicit the best outputs achievable from visual AR models.
    \item We propose a layout-specified prompt reformulation that leverages partial views to infer feasible layouts, tackling the blueprint deficiency in autoregressive generation and enriching the candidate pool with prompt-aligned images for more effective test-time scaling.
    \item  Extensive experiments demonstrate that \textit{GridAR} improves generation quality across both text-to-image generation and image editing, drawing out the maximum potential of the pretrained visual AR model and offering a superior cost-quality trade-off.
\end{itemize}

\section{Preliminary}
\subsection{Autoregressive Modeling for Image Generation}
Visual autoregressive (AR) models adapt the next-token prediction paradigm of language modeling to image generation. To enable autoregressive prediction on images, they employ a vector-quantized autoencoder, such as VQ-VAE~\citep{van2017neural} and VQ-GAN~\citep{esser2021taming}, which discretizes an image into a finite sequence of codebook indices. Given a trained vector-quantized autoencoder with down-sampling factor $M$ and codebook $\mathcal{Q}=\{\ve_k\}_{k=1}^K$, an image $\mI \in \sR^{H \times W \times 3}$ is encoded into a grid of $h \times w$ latent vectors, where $h = H/M$ and $w = W/M$. 
These latent vectors are then quantized by the codebook into a discrete sequence $\vx = (x_1,\ldots,x_N)$, where $x_i \in \{1,2,\dots, K\}$ and $N = h \cdot w$. 

In text-to-image generation or image editing, the image token sequence $\vx$ is generated conditioned on context $\vc$ (e.g., text or image embeddings). AR models perform next-token prediction on $\vx$, defining the sequence likelihood as:
\vspace{-0.75em}
\begin{equation*}
    p\big(\vx\mid \vc\big)=\prod_{n=1}^{N} p\big(x_n \mid x_{<n},\, \vc\big).
\end{equation*}

For a visual AR model $p_\phi(\vx\mid\vc)$ conditioned on context $\vc$, training updates the parameters $\phi$ so that $p_\phi(x_n \mid x_{<n},\, \vc)$ fits the dataset distribution. During inference, the model samples tokens autoregressively, after which a decoder reconstructs the high-resolution image from the corresponding codebook embeddings. Related work on visual AR models is also discussed in Appendix \textcolor{red}{A}.
\vspace{-0.01in}
\subsection{Classifier-Free Guidance for AR Models}

Classifier-free guidance (CFG)~\citep{cfg} was first introduced for diffusion models to eliminate external classifiers and control a trade-off between fidelity and conditional adherence. At each sampling step, the score estimate is formed as a linear combination of the conditional and unconditional scores.
Early intuition suggested that CFG draws samples from a reweighted distribution proportional to $p(\vx\mid \vc)^{s+1}  p(\vx)^{-s}$~\citep{cfg} where $s$ is the guidance scale; however, later analyses show this interpretation is generally incorrect and is better viewed as a predictor-corrector procedure~\citep{bradley2024classifier}.

CFG has since been adapted to diverse AR models-such as text-to-image~\citep{llamagen, wang2024emu3, janus, januspro, liu2024lumina} and text-to-music~\citep{copet2023simple}-through logit-space guidance applied before next-token sampling. For an AR model $p_\phi(\vx\mid \vc)$, to generate $x_i$ at step $i \in{\{1, 2, \dots, N\}}$, the guided logits $l^{\text{sample}}_i$ are formed as follows: 
\vspace{-0.35em}
\begin{equation*}
l_i^{\text{sample}} = (1+s) \cdot l^{\text{cond}}_i - s \cdot l^{\text{uncond}}_i = l^{\text{cond}}_i + s \cdot \big(l^{\text{cond}}_i - l^{\text{uncond}}_i\big),
\end{equation*}
where $l^{\text{cond}}_i$ and $l^{\text{uncond}}_i$ are the conditional and unconditional logits produced by the same model. The next-token distribution can be represented as:
\vspace{-0.35em}
\begin{equation*}
\begin{aligned}
p^{\text{sample}}_\phi\big(x_i\mid x_{<i},\, \vc\big)&=\softmax(l_i^{\text{sample}}) \\
&\propto p_\phi\big(x_i\mid x_{<i},\, \vc\big) \left(\frac{p_\phi\big(x_i\mid x_{<i},\, \vc\big)}{p_\phi\big(x_i\mid x_{<i}\big)}\right)^s.
\end{aligned}
\end{equation*}
Using the autoregressive modeling on $\vx$, the induced sequence-level sampling distribution satisfies
\vspace{-0.35em}
\begin{equation*}
p^{\text{sample}}_\phi\big(\vx\mid \vc\big) \propto p_\phi\big(\vx\mid \vc\big) \left(\frac{p_\phi\big(\vx\mid \vc\big)}{p_\phi\big(\vx\big)}\right)^s.
\end{equation*}

\section{Test-Time Scaling for Autoregressive Image Generation}~\label{sec:method}
\begin{figure*}[t]
    \vspace{-1.5em}
    \centering
    \includegraphics[width=0.93\linewidth]{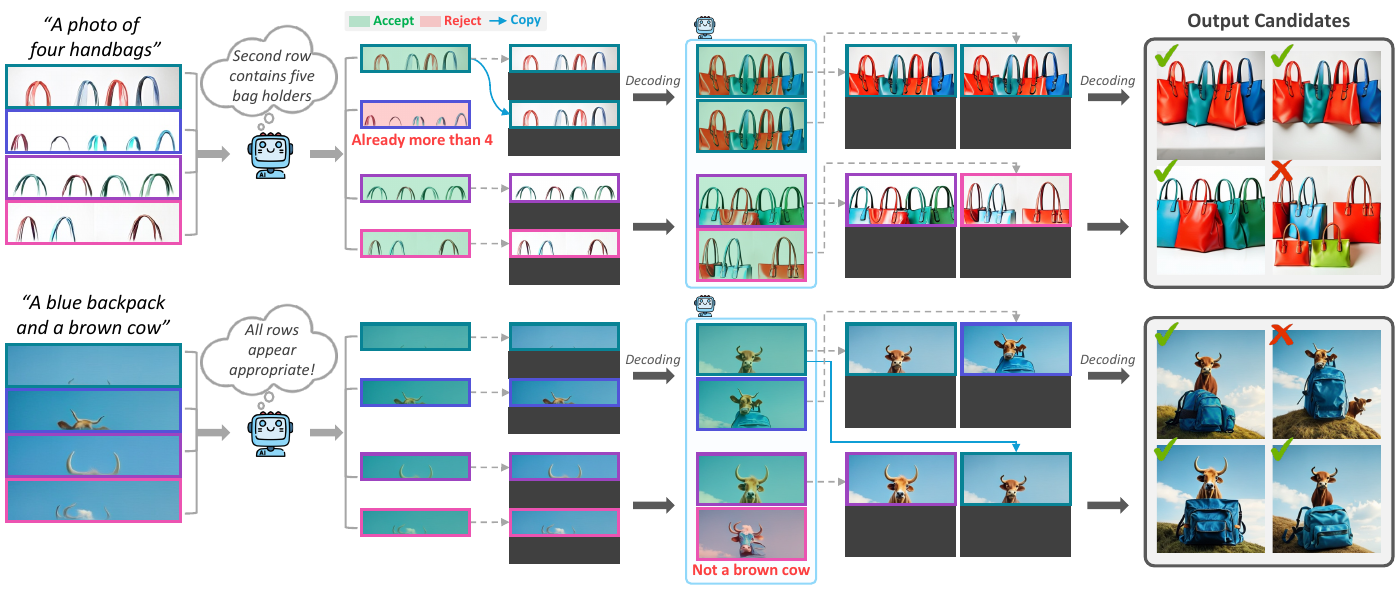}
    \vspace{-1em}
    \caption{\textbf{Visualization of Grid-based Progressive Generation} process in two cases: (a) first-stage rejection (top row), where all candidates are accepted in the second stage; (b) second-stage rejection (bottom row), where all candidates are accepted in the first stage.}
    \vspace{-1.5em}
    \label{fig:grid_verification}
\end{figure*}

Our main objective is to investigate how test-time computation can be scaled to elicit the best outputs from visual autoregressive (AR) models with next-token prediction. We introduce \textbf{\textit{GridAR}}, a test-time scaling framework that progressively explores generation paths through grid-structured canvases, dynamically directing computation toward promising continuations. Multiple partial candidates for the same position are generated in a row-wise form; unlikely ones are pruned early, and viable ones are fixed as anchors to guide subsequent generation (Section~\ref{subsec:grid_generation}). We also incorporate a layout-aware prompt reformulation alongside candidate verification (Section~\ref{subsec:prompt_reformulation}). By restructuring the prompt to reflect a feasible layout consistent with selected intermediate results, this step addresses the blueprint deficiency of raster-scan decoding. The reformulated prompt is then used in subsequent decoding  - either with three-way classifier-free guidance or simple prompt replacement - to ensure the remaining regions. These two components strengthen the candidate pool, leading to more reliable text-to-image generation. An overview is illustrated in Figure~\ref{fig:grid_verification}. 

\subsection{Grid-Based Progressive Generation}~\label{subsec:grid_generation}
In this section, we describe our progressive image completion process through parallel candidate exploration, during which promising candidates are retained - a sophisticated version of Best-of-$N$ for visual AR models. We employ a row-partitioned generation strategy: the first stage uses an $R_{1}$-row grid to explore $R_{1}$ initial partial candidates, followed by an $R_{2}$-row grid for continued generation from the selected anchor candidates. We instantiate $(R_{1}, R_{2}) = (4,2)$ here as it balances information and computation; while other configurations are possible and also generalize, as analyzed in Appendix \textcolor{red}{D}.

Starting from this setup, the text-to-image autoregressive model $p_{\phi}(\vx\mid\vc_{T})$ begins by generating four distinct candidates ($R_{1}= 4$), each corresponding to the upper quarter of the image given the same prompt. Specifically, we represent the canvas $\vx \in \{1, 2, \dots, K\}^{h \times w}$ as four contiguous horizontal row segments, $\vx = \left[\begin{smallmatrix} \vx^{(1)}\\ \vdots\\ \vx^{(4)} \end{smallmatrix}\right]$, where $\vx^{(r)}$ is the $r$-th row segment and each segment consists of L tokens with $L=\tfrac{h}{4}\cdot w$. Under this partition, each candidate is autoregressively generated as:
\vspace{-0.1in}
\begin{equation*}
p_\phi\!\big(\vx^{(r)} \mid \vc_{T}\big)
=\prod_{n=1}^{L} p_\phi\!\left(x^{(r)}_n \,\middle|\, x^{(r)}_{<n},\, \vc_{T}\right),
\quad r=1,\dots,4,
\end{equation*}
where $x^{(r)}_{n}$ is the $n$-th discrete token index. Different candidates $\vx^{(r)}$ are generated independently (\ie without conditioning on each other), while the key-value representations of the prompt $\vc_{T}$ are cached once and reused across all rows for efficiency. Then grid-partitioned canvas $\mI_{\text{grid}}$ containing four candidates is decoded from the vector-quantized embeddings $\vx^{q} \in \mathbb{R}^{h \times w \times d}$ (obtained by mapping $\vx$ to its codebook vectors) through a single forward pass of the decoder $\mathcal{D}_{\text{VQ}} : \mathbb{R}^{h \times w \times d} \to \mathbb{R}^{h \times w \times 3}$ as $\mI_{\text{grid}}=\mathcal{D}^{\text{VQ}}\big(\vx^q\big)$.

\vspace{-0.07in}
\paragraph{Candidate verification} After obtaining image $\mI_{\text{grid}}$, we assess the four candidate row-segment images \textit{at once} using a verifier $V_{\psi}$. We here employ a vision-language model as a zero-shot verifier to determine whether candidates are already unlikely to satisfy the given prompt - \eg, when attribute bindings such as color are already incorrect, or when the number of objects exceeds what is required. The verifier $V_{\psi}$ predicts row-wise judgments directly as:
\vspace{-0.75em}
\begin{equation*}
\begin{aligned}
\mathbf{y} = V_\psi\big(\mI_{\text{grid}},\,\vc_{T}\big)
= \big(y^{(1)},\dots,y^{(4)}\big),\\
y^{(r)}\in\{\texttt{possible},~\texttt{impossible}\}.
\end{aligned}
\end{equation*}

It is worth noting that we do not perform top-$k$ selection as in beam search~\citep{sutskever2014sequence}; instead, we reject only those candidates deemed impossible to align with the prompt, while keeping all others. In other words, the number of candidates retained is not fixed~\footnote{In rare cases, all candidates may be rejected; the frequency of such events and our handling strategy are detailed in Appendix \textcolor{red}{C}}. The intuition behind this is that we are evaluating only partial views, where a certain object may simply not appear yet in the visible rows. As a result, top-$k$ selection may prematurely discard many potentially valid candidates for such reasons, thereby harming the sample diversity of the candidate pool. 
According to the verified results, the set of tokens for four anchor partial images is determined, which will serve as anchors to guide the continued generation. If some candidates are rejected, the rejected one is randomly replaced with feasible one, for example, if $\vx^{(2)}$ rejected, the anchor set  $\vx_{\text{anchor}}$ can be determined as $\vx_{\text{anchor}}$ =  $[\,\vx^{(1)}, \vx^{(4)}, \vx^{(3)}, \vx^{(4)}\,]^{\top}$. As we employ a zero-shot verifier, analyzing its accuracy and its effect is essential; we examine this in Appendix C.

\begin{figure*}[t]
    \vspace{-1em}
    \centering
    \includegraphics[width=0.98\linewidth]{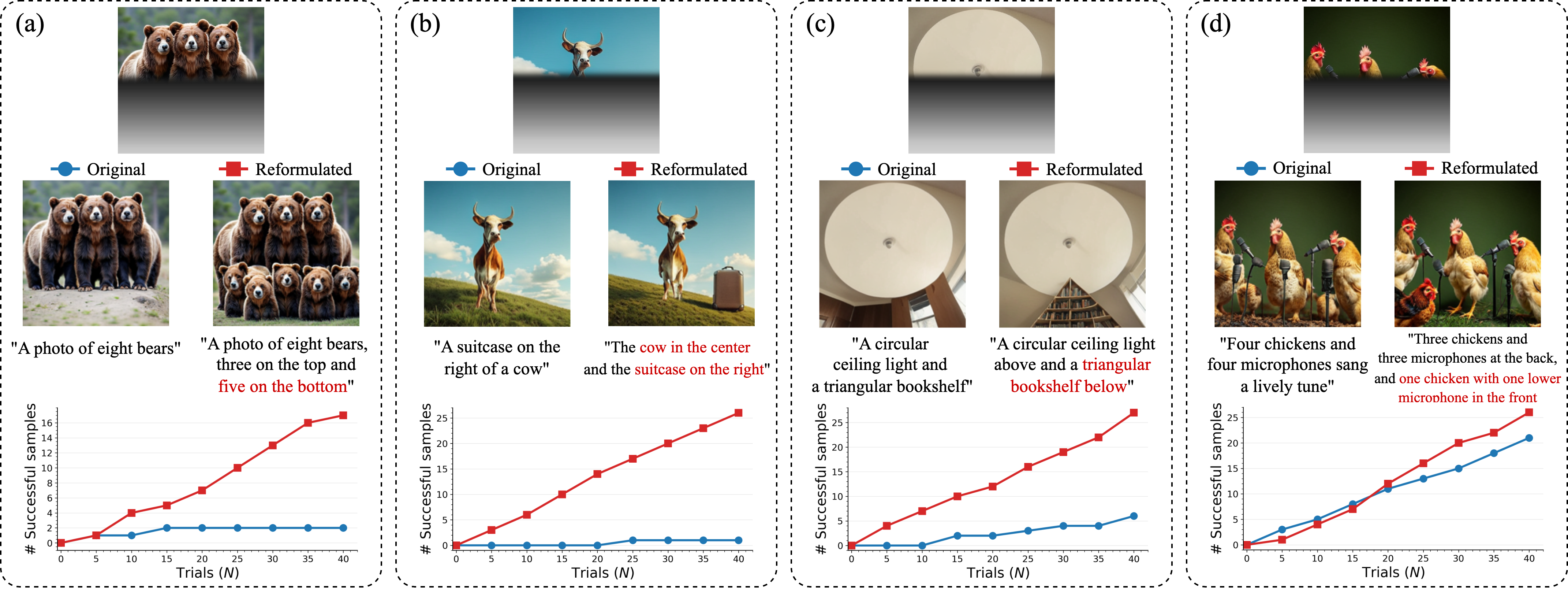}
    \vspace{-0.5em}
    \caption{\textbf{Motivation of Prompt Reformulation.} Success rate increases significantly with the number of trials when prompt reformulation incorporates a plan for generating lower tokens, rather than relying only on the tokens generated in the upper part.}
    \vspace{-0.9em}
    \label{fig:prompt_reformulation}
\end{figure*}

\paragraph{Image expansion from verified candidates} 
The discrete token sequences of four verified partial images $\{\vx_{\text{anchor}}^{(r)}\}_{r=1}^{4}$ are propagated to the next stage, where two distinct grids with $R_{2}$ rows ($R_{2}$ = 2). In each grid cell, the upper portion is fixed by anchor tokens, and the lower portion is autoregressively continued. Under $(R_{1},R_{2})$ = $(4,2)$, this yields two half-image canvases guided by their anchors: 
$\tilde{\vx}_{1}$ = $[\,(\vx_{\text{anchor}}^{(1)},\,\vx_{\text{gen}}^{(1)}),\,(\vx_{\text{anchor}}^{(2)},\,\vx_{\text{gen}}^{(2)})\,]^{\top}$ 
and 
$\tilde{\vx}_{2}$ = $[\,(\vx_{\text{anchor}}^{(3)},\,\vx_{\text{gen}}^{(3)}),\,(\vx_{\text{anchor}}^{(4)},\,\vx_{\text{gen}}^{(4)})\,]^{\top}$, 
where each pair consists of a fixed anchor and its autoregressive continuation, 
$p_\phi(\vx_{\text{gen}}^{(i)}\mid \vx_{\text{anchor}}^{(i)},\vc_{T})$=$\prod_{n=1}^{L'}p_\phi(x_{\text{gen},n}^{(i)}\mid x_{\text{gen},<n}^{(i)},\vx_{\text{anchor}}^{(i)},\vc_{T})$ with $L'$ denoting the number of tokens in the half-image segment.

This process yields four half-image views arranged in a 2-by-2 layout. As before, the verifier $V_{\psi}$ prunes half-image candidates unlikely to satisfy the prompt, with rejected candidates substituted by viable ones. Each verified half-image is then anchored on a new canvas, and the remaining half autoregressively generated to form four complete images. The best image is selected using an output reward model (ORM), analogous to Best-of-$N$. Apart from verification overhead, the number of tokens matches that of $N$=4 in standard Best-of-$N$. Although we describe the $N$=4 case for clarity, the framework scales naturally,- \eg, two starting canvases yield $N$=8. Further analysis of rejection rates and rejected samples is provided in Appendix \textcolor{red}{B}.

\subsection{Layout-Specified Prompt Reformulation}~\label{subsec:prompt_reformulation}
As described in Section~\ref{subsec:grid_generation}, our grid-based image generation framework effectively enlarges the search space while circumventing wasted computation on erroneous trajectories. Nevertheless, this pipeline alone is insufficient to ensure a strong candidate pool. Even with carefully selected anchors in the upper half, subsequent decoding may still repeat objects already drawn or omit others required by the prompt. We posit that such failures arise from the absence of a global blueprint: due to the next-token prediction nature of auto-regressive decoding, the model lacks an explicit plan for how the prompt should be realized across the entire canvas. To probe this limitation, we conduct a pilot study that verifies blueprint deficiency as one of the bottlenecks in faithfully portraying prompts within visual AR models. This motivates our \textit{layout-specified prompt reformulation}, which dynamically revises the prompt using plausible layouts inferred from the intermediate canvases.

\begin{table*}[t]
\vspace{-1.5em}
\caption{T2I-CompBench++ results by dimensions, comparing Best-of-\textit{N} with \textit{GridAR} test-time scaling on Janus-Pro and LlamaGen. Scores are reported using metrics proposed in the benchmark.}
\vspace{-1.75em}
\begin{center}
\begin{small}
\setlength{\columnsep}{1.0pt}%
\begin{adjustbox}{width=0.9\linewidth}
\begin{tabular}{lcccccccc}
\toprule
\multirow{2}{*}{\textbf{Method}} &  \multicolumn{3}{c}{\textbf{Attribute Binding}} &  \multicolumn{3}{c}{\textbf{Object Relationship}} &  \multirow{2}{*}{\textbf{Numeracy}}  & \multirow{2}{*}{\textbf{Complex}} \\
\cline{2-4} \cline{5-7}
& Color & Shape & Texture & 2D Spatial & 3D Spatial & Non-Spatial & &  \\
\cline{1-9}
\rowcolor{lightgray}\textit{Diffusion Models } & & & & & & & & \\
SDXL & 0.5879 & 0.4687 & 0.5299 & 0.2133 & 0.3566 & 0.7673 & 0.4988 & 0.3237 \\
Pixart-$\alpha$ & 0.6690 & 0.4927 & 0.6477 & 0.2064 & 0.3901 & 0.7747 & 0.5032 & 0.3433 \\
DALL$\cdot$E~3 & 0.7785 & 0.6205 & 0.7036 & 0.2865 & 0.3744 & 0.7853 & 0.5926 & 0.3773 \\
SD3 & 0.8132 & 0.5885 & 0.7334 & 0.3200 & 0.4084 & 0.7782 & 0.6174 & 0.3771 \\
FLUX.1 & 0.7407 & 0.5718 & 0.6922 & 0.2863 & 0.3866 & 0.7809 & 0.6185 & 0.3703 \\ 
                    
\rowcolor{lightgray} \textit{Auto-Regressive Models} & & & & & & & &\\
Lumina-mGPT & 0.6371 & 0.4727 & 0.6034 & - & - & - & - & -  \\
Emu3 & 0.6107 & 0.4734 & 0.6178 & - & - & - & - & -  \\
\cdashline{1-9}
LlamaGen & 0.2927 & 0.3160 & 0.3828 & 0.1118 & 0.1510 & 0.7143 & 0.2727 & 0.2445 \\
LlamaGen + BoN (N=8) & \underline{0.5143} & 0.4465 & \underline{0.5850} & \underline{0.1578} & \underline{0.2016} & \underline{0.7543} & \underline{0.4197} & \underline{0.3054} \\
\rowcolor{lightblue}\textbf{LlamaGen + GridAR (N=4)} & 0.4969 & \underline{0.4540} & 0.5675 & 0.1466 & 0.1946 & 0.7470 & 0.4118 & 0.2993\\
\rowcolor{lightblue}\textbf{LlamaGen + GridAR (N=8)} & \textbf{0.5774} & \textbf{0.4783} & \textbf{0.5984} & \textbf{0.1830} & \textbf{0.2019} & \textbf{0.7570} & \textbf{0.4407} & \textbf{0.3103}\\
\cdashline{1-9}
Janus-Pro & 0.5388 & 0.3476 & 0.4357 & 0.1607 & 0.2806 & 0.7733 & 0.4467 & 0.3796\\
Janus-Pro + BoN (N=8) & 0.7234 & 0.4178 & 0.5600 & 0.2430 & 0.3165 & 0.7853 & 0.5068 & \underline{0.3926} \\
\rowcolor{lightblue}\textbf{Janus-Pro + GridAR (N=4)} & \underline{0.8050} & \underline{0.6014} & \underline{0.7268} & \underline{0.2833} & \underline{0.3503} & \underline{0.7887} & \underline{0.5684} & 0.3905\\
\rowcolor{lightblue}\textbf{Janus-Pro + GridAR (N=8)} & \textbf{0.8172} & \textbf{0.6174} & \textbf{0.7408} & \textbf{0.3214} & \textbf{0.3587} & \textbf{0.7930} & \textbf{0.5932} & \textbf{0.4041}\\

\bottomrule
\end{tabular}
\end{adjustbox}
\end{small}
\end{center}
\label{tb:main_t2icompbench++}
\vspace{-2.3em}
\end{table*}
\vspace{-0.16in}
\paragraph{Pilot study} We test whether raster-scan decoding, which generates tokens sequentially without awareness of the overall layout, can construct a high-quality candidate pool under test-time scaling, and whether injecting layout knowledge mid-generation can remedy this limitation. Using Janus-Pro 7B~\citep{janus}, we focus on prompts where the single-sample setting ($N$=1) fails. As shown in Figure~\ref{fig:prompt_reformulation}, we analyze partially decoded images that remain correct up to the halfway point, conditioning on the upper-half tokens and repeatedly decoding the lower half to measure how many samples eventually satisfy the prompt as trials increase. Results show that many candidates remain incorrect - often duplicating or omitting objects - even with Best-of-$N$ selection, despite the prompt being achievable (blue curves). Instead of persisting with the original prompt, we reformulate it at the intermediate stage, replacing it with a revised version for subsequent decoding.
We revise the prompt by inspecting the intermediate output and specifying a feasible layout that can satisfy the prompt. For instance, in  Figure~\ref{fig:prompt_reformulation} (a), the prompt \textit{“a photo of eight bears”} is reformulated after the partial grid already depicts three bears in the upper region,  yielding \textit{“three on top and five on the bottom”}. This simple modification leads to a clear improvement in candidate quality, as shown by the consistently higher success rates as scaling increases  (red curves). These results indicate that prompt reformulation allows strong candidate pools to be obtained even with lower levels of scaling. 
\vspace{-0.12in}
\paragraph{Prompt reformulation} Motivated by this study, we incorporate prompt reformulation into the grid-based progressive generation process. When the verifier $\psi$ evaluates grid candidates, we simultaneously conduct a \textit{layout-specified prompt reformulation}, in which the original prompt is revised to reflect a realizable layout consistent with the observed partial images. The reformulated prompt provides explicit structural cues (\eg, object count or spatial arrangement) inferred from the verified candidates. We consider two alternative strategies: (i) a three-way classifier-free guidance (CFG) that steers logits toward the specified layout by orthogonalizing the reformulated prompt against the original, and (ii) a cost-efficient approach that simply replaces the prompt in subsequent decoding. 
\vspace{-0.12in}
\paragraph{(i) Three-way CFG} 
Let $T_{u}$, $T_{o}$, and $T_{r}$ denote unconditional (null text), original, and reformulated prompts, respectively. At token step $i$, the autoregressive model $f_{\theta}$ produces a hidden representation, which is projected by the generation head $W$ into the logit space as: 
$l^{(u)}_i = W f_{\theta}(x_{<i}, i, T_{u}), \;
 l^{(o)}_i = W f_{\theta}(x_{<i}, i, T_{o}), \;
 l^{(r)}_i = W f_{\theta}(x_{<i}, i, T_{r}).$ 
We then derive two directional offsets as
$d_{o,i} = l^{(o)}_i - l^{(u)}_i, \;
 d_{r,i} = l^{(r)}_i - l^{(u)}_i.$ 
To disentangle the layout-specific direction from the original one, we orthogonalize $d_{r,i}$ against $d_{o,i}$, ensuring no interference with the original guidance scale while clearly conveying the layout-specific direction: $\tilde{d}_{r,i} = d_{r,i} - \frac{\langle d_{r,i}, d_{o,i}\rangle}{\|d_{o,i}\|^2} \, d_{o,i}.$
Finally, the three-way CFG logits are defined as: $l_i^{\text{sample}} = l^{(o)}_i +s_{o} \cdot d_{o,i} + s_{r} \cdot \tilde{d}_{r,i}$,
where $s_{o}, s_{r} \geq 0$ are guidance scales controlling the strengths of the original and reformulated signals. In this work, we do not tune these parameters but simply set $s_o= s_r = s$ to a fixed constant (reusing the conventional scale $s$ for both scales). This formulation preserves the contribution of the original prompt while providing a clear layout-specific signal from the reformulated prompt.
\vspace{-0.13in}
\paragraph{(ii) Prompt replacement}
As a cost-efficient alternative, we directly substitute $T_{r}$ for $T_{o}$ in subsequent decoding steps without modifying the logit computation of classifier-free guidance. Although this strategy does not match the fine-grained signal of a three-way CFG, it offers a lightweight option that still decently guides the model toward layouts consistent with the verified intermediate results. The logit $l_i^{\text{sample}}$ under this strategy follows the standard classifier-free guidance formulation: $l_i^{\text{sample}} = l^{(r)}_i + s_{r} \cdot {d}_{r,i} = l^{(r)}_i + s_{r} \cdot (l^{(r)}_i - l^{(u)}_i) .$
The same CFG scale $s_r$ used in the earlier image generation with the original prompt is reused here. We show that this strategy outperforms approaches that employ a planner to specify layouts prior to generation in AR models (see Section~\ref{subsec:exp_analysis}).

\vspace{-0.07in}
\section{Experiments}~\label{sec:experiments}
We conduct a collection of experiments to validate \textit{GridAR} on visual autoregressive (AR) models through two primary tasks: text-to-image generation and image editing. First, we show that our framework can consistently elicit superior image generation results compared to existing test-time scaling methods across diverse prompt categories (Section~\ref{subsec:exp_t2i}). We then demonstrate its versatility in image editing, where the model receives both an edit instruction and a source image, and verify that \textit{GridAR} likewise enhances the effectiveness of computation scaling for this task (Section~\ref{subsec:exp_edit}). Beyond these benchmark evaluations, we further conduct in-depth analyses addressing research questions raised by our framework, including robustness to different verifiers and comparisons between design choices (Section~\ref{subsec:exp_analysis}). Lastly, we conduct a \textbf{human evaluation} and present an \textbf{ablation study} along with a comparison of two prompt-reformulation strategies (Appendix \textcolor{red}{D}).

\subsection{Experimental Setup}
\paragraph{Implementation details}
We use Janus-Pro-7B~\citep{janus} and LlamaGen~\citep{llamagen} as backbones for autoregressive text-to-image generation, and EditAR~\citep{mu2025editAR} for image editing tasks. Across all experiments, Qwen2.5-VL~\citep{bai2025qwen2} is employed as the outcome reward model for both our method and test-time scaling baselines. For the CFG scale, we set $s_o=5$ for Janus-Pro, and $s_o=6.5$ for LlamaGen, following the original paper setup. Three-way CFG is used as default. In \textit{GridAR}, the guidance scale for the reformulated prompt is set equal to the original ($s_r=s_o$). To evaluate candidates and conduct prompt reformulations, we deploy GPT-4.1 as the verifier $V_{\psi}$, while other verifiers are tested in Section~\ref{subsec:exp_analysis}.

\begin{figure*}[t]
    \vspace{-1.5em}
    \centering
    \includegraphics[width=1.\linewidth]{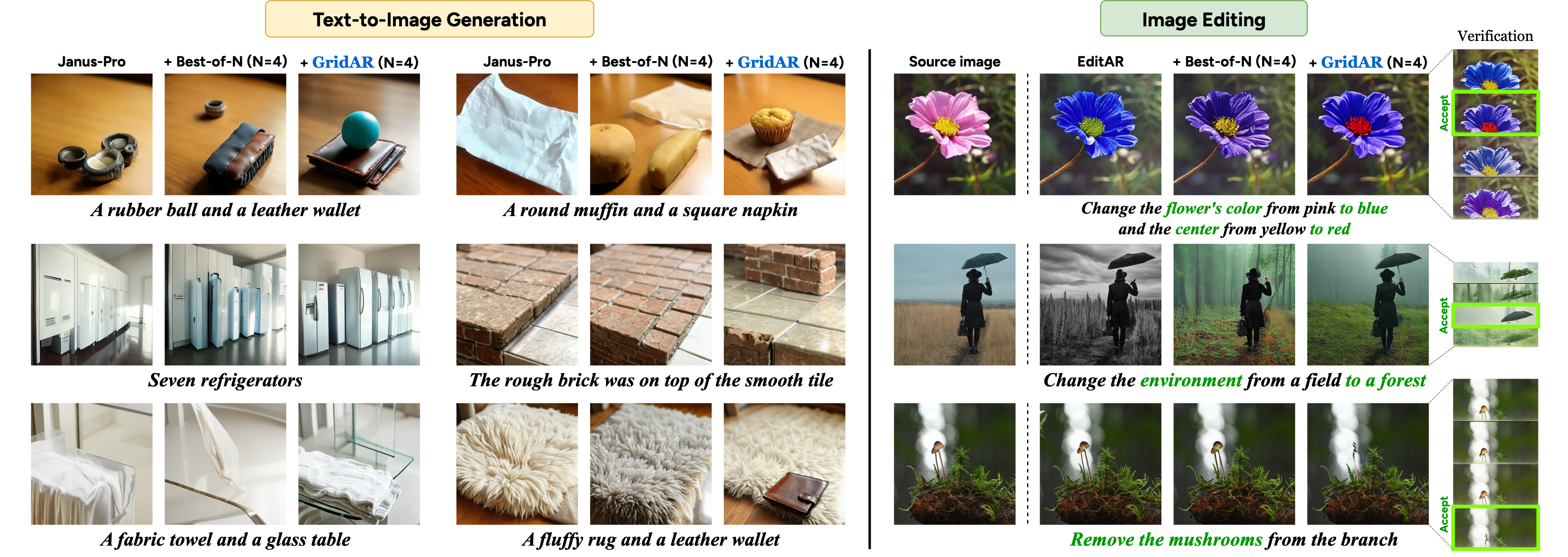}
    
    \vspace{-0.5em}
    \caption{\textbf{Qualitative Results} comparing single-generation outputs, Best-of-$N$ $(N=4)$ outputs, and outputs obtained by applying \textit{\textbf{GridAR}} $(N=4)$ on text-to-image generation and image editing.}
    \label{fig:qualitative_main}
    \vspace{-1.5em}
\end{figure*}

\vspace{-0.5em}
\paragraph{Datasets and metrics}
Text-to-image generation is evaluated on two benchmarks: T2I-CompBench++~\citep{huang2025t2i} and GenEval~\citep{ghosh2023geneval}. T2I-CompBench++ comprises 8,000 compositional prompts across seven categories. Evaluation follows the metrics proposed in the original paper: BLIP-VQA (Color, Shape, Texture), UniDet (2D/3D Spatial, Numeracy), ShareGPT4V (Non-Spatial), and the 3-in-1 score (Complex). GenEval includes over 500 prompts from six categories, and performance is measured by binary correctness on compositional properties, using models such as Mask2Former~\citep{cheng2022masked} and CLIP ViT-L/14~\citep{radford2021learning}. For image editing, we evaluate on PIE-Bench~\citep{PnP_Inversion_ju2023direct}, covering 9 editing scenarios and 700 images with paired source images and edit instructions. Performance is assessed along two axes: instruction following and semantic preservation. Instruction following is measured by the CLIP similarity~\citep{hessel2021clipscore}, and source preservation is measured by structure distance with DINO-ViT~\citep{caron2021emerging} and various perceptual metrics - PSNR, LPIPS~\citep{zhang2018unreasonable}, MSE, and SSIM~\citep{wang2004image}. We primarily compared our method against Best-of-\textit{N} scaling, and include AR and diffusion-based models as baselines. Evaluation protocols and baseline details are provided in Appendix \textcolor{red}{F}.

\begin{table}[t]
\centering
\scriptsize
\setlength{\tabcolsep}{2pt}
\renewcommand{\arraystretch}{0.95}

\caption{GenEval results on three selected dimensions and overall.}
\vspace{-0.75em}
\label{tb:main_geneval}
\begin{adjustbox}{width=0.95\linewidth}
\begin{tabular}{lcccc}
\toprule
\multirow{2}{*}{\textbf{Method}} & \multirow{2}{*}{Counting} &  \multirow{2}{*}{Position} &  \multirow{2}{*}{\shortstack[1]{Color \\ Attribution}} & \multirow{2}{*}{\textbf{Overall}}  \\
\\
\cline{1-5} 
\rowcolor{lightgray}\textit{Diffusion Models } & & & &   \\
SDXL &   0.39 &  0.15 & 0.23 & 0.55 \\
Pixart-$\alpha$ & 0.44  & 0.08 & 0.07 & 0.48 \\
DALL$\cdot$E~3 & 0.47  & 0.43 & 0.45 & 0.67 \\
SD3 & 0.72 & 0.33 & 0.60 & 0.74 \\
\rowcolor{lightgray} \textit{Auto-Regressive Models} & & & &  \\
Show-o & 0.49  & 0.11 & 0.28 & 0.53 \\
Show-o + PARM (N=20) & 0.68 & 0.29 & 0.45 & 0.67\\
Emu3 & 0.34 & 0.17 & 0.21 & 0.54 \\
Infinity & - & 0.49 & 0.57 & 0.73\\
\cdashline{1-5}
LlamaGen  & 0.12 & 0.14 & 0.05 & 0.34\\
LlamaGen + BoN (N=8) & \underline{0.21}  & \underline{0.22} & \textbf{0.14} & \underline{0.44} \\
\rowcolor{lightblue}\textbf{LlamaGen + GridAR (N=8)}  & \textbf{0.24}  & \textbf{0.25} & \underline{0.13} & \textbf{0.46} \\
\cdashline{1-5}
Janus-Pro  & 0.59  & 0.77 & 0.65 & 0.79 \\
Janus-Pro + BoN (N=8)  & \underline{0.76} & \underline{0.86} & \underline{0.72} & \underline{0.86} \\
\rowcolor{lightblue}\textbf{Janus-Pro + GridAR (N=8)} & \textbf{0.79}  & \textbf{0.92} & \textbf{0.73} & \textbf{0.88} \\
\bottomrule
\end{tabular}
\end{adjustbox}
\vspace{-1.5em}
\end{table}

\subsection{Text-to-Image Generation}~\label{subsec:exp_t2i}
We test the effectiveness of \textit{GridAR} in improving image generation quality by scaling test-time compute on text-to-image benchmarks. As shown in Table~\ref{tb:main_t2icompbench++}, \textit{GridAR} improves the average score by 17.8\% and 4.8\% for Janus-Pro and LlamaGen, respectively, under the same \(N\) across diverse prompt scenarios. Notably, \textit{GridAR} with \(N{=}4\) even outperforms Best-of-\(N{=}8\) on Janus-Pro, achieving a gain of 14.4\% and demonstrating a better cost-performance trade-off (see Section~\ref{subsec:exp_analysis}). These results suggest that our framework is able to derive a higher-quality candidate pool. In particular, when paired with stronger visual AR models such as Janus-Pro, the synergy appears more pronounced-likely due to their improved ability to follow layout specifications and to generate more accurate initial candidates. For the GenEval benchmark, we report performance on Counting, Position, and Color Attribution tasks in Table~\ref{tb:main_geneval} (other dimensions are already saturated at scores near 0.90-0.99; overall results are provided in the last column). Our method also enhances text-to-image generation quality over Best-of-\(N\) across most dimensions. 
We also compare \textit{GridAR} with test-time scaling methods~\citep{chen2025ttsvar, t2ir1} from other AR families and those based on reinforcement learning, with results in Appendix \textcolor{red}{D}. In addition, we provide qualitative samples in Figure~\ref{fig:qualitative_main} (left) and Appendix \textcolor{red}{G}, showing how \textit{GridAR} leads to more accurate instruction-aligned final selections.

\begin{figure}
    \vspace{-0.8em}
    \begin{subfigure}[t]{0.49\linewidth}
        \centering
        \includegraphics[width=\linewidth]{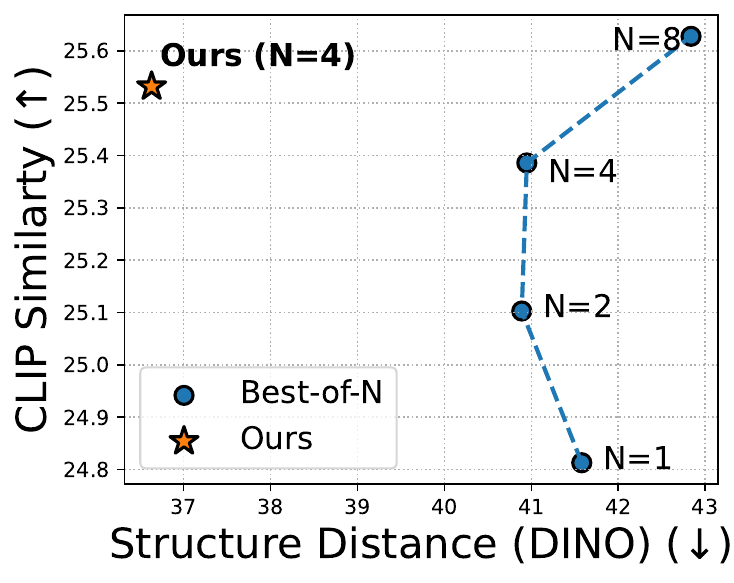}
    \end{subfigure}
    \begin{subfigure}[t]{0.49\linewidth}
        \centering
        \includegraphics[width=\linewidth]{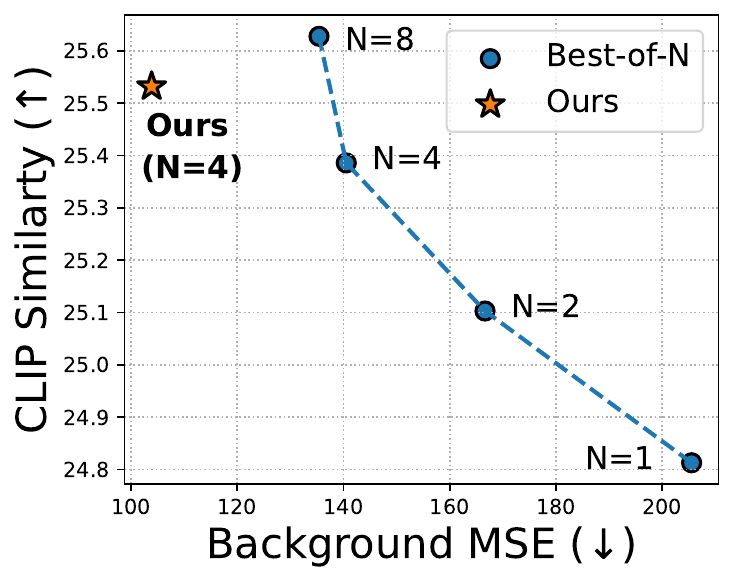}
    \end{subfigure}
    \vspace{-0.5em}
    \caption{Image editing results on PIE-Bench.} 
    \vspace{-1.8em}
    \label{fig:editing_result_plot}
\end{figure}

\begin{figure*}[t]
    \centering
    \vspace{-1em}
    \begin{minipage}{0.29\linewidth}
        \centering
        \includegraphics[width=\linewidth]{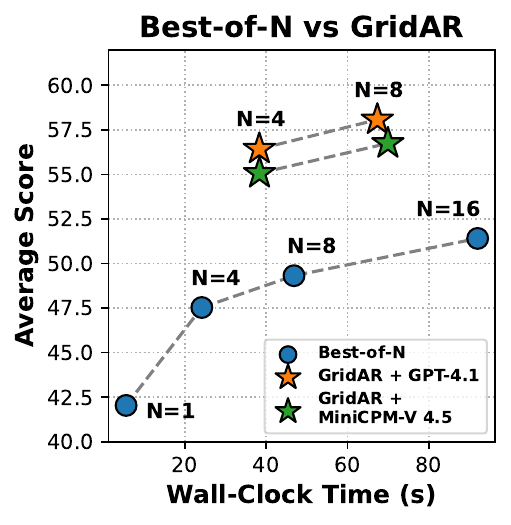}
    \end{minipage}
    \hfill
    \begin{minipage}{0.30\linewidth}
        \centering
        \includegraphics[width=\linewidth]{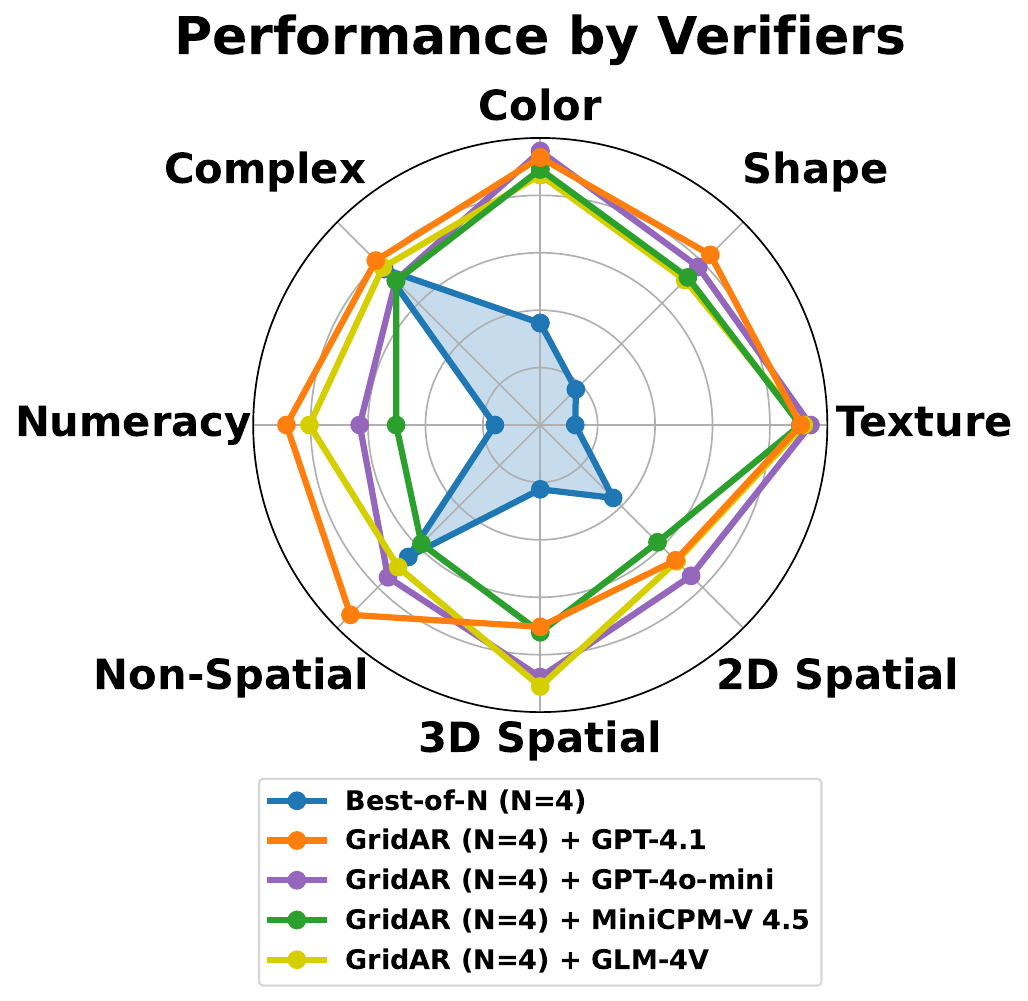}
    \end{minipage}
    \hfill
    \begin{minipage}{0.30\linewidth}
        \centering
        \vspace{-1.75em}
        \includegraphics[width=\linewidth]{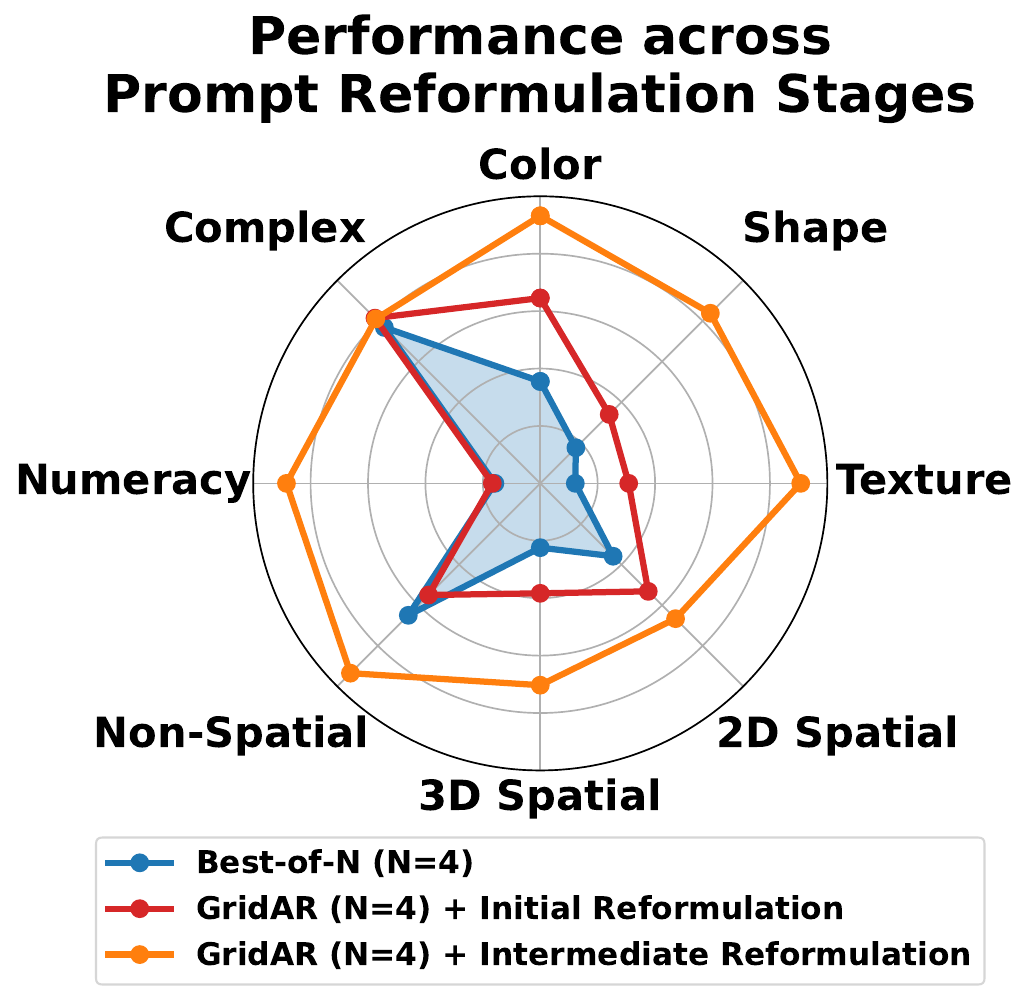}
    \end{minipage}
    \vspace{-1em}
    \caption{\textbf{Left}: Computation-performance trade-off; \textbf{Center}: Performance across dimensions by different verifiers; \textbf{Right}: Performance across dimensions by different prompt reformulation timings.} 
    \vspace{-1.3em}
    \label{fig:in_depth_anlysis}
\end{figure*}

\subsection{Results on Image Editing}~\label{subsec:exp_edit}
We validate that our test-time scaling framework can be extended naturally to image editing and boost the quality of edited images. To adapt \textit{GridAR} to this setting, we modify the prompt for the verifier to account for both source preservation and adherence to edit instructions. Prompt reformulation is applied similarly to text-to-image generation, where feasible layouts are inferred from intermediate results to satisfy the edit instruction.  We compare our method against Best-of-\textit{N} scaling using the same backbone (EditAR), as well as test-time scaled diffusion-based editing models, across 7 metrics. As shown in Figure~\ref{fig:editing_result_plot} and Appendix \textcolor{red}{D.2}, \textit{GridAR} significantly improves background preservation over Best-of-\textit{N} ($N$=4), reducing structure-aware distance by $7.27$\% and MSE by $17.01$\%. Edit instruction fidelity, measured by CLIP similarity, also shows improvement.  Even when compared to Best-of-\textit{N} ($N$=8), our method achieves comparable CLIP similarity (25.532 vs. 25.628), while substantially outperforming in source preservation - showing lower structure-aware distance (36.632 vs. 42.873) and lower MSE (103.896 vs. 135.404). Edited samples are provided in Figure~\ref{fig:qualitative_main} (right), and full quantitative results, including comparisons with test-time scaled diffusion-based models, are reported in Appendix \textcolor{red}{D.2}.


\subsection{In-depth Analysis}~\label{subsec:exp_analysis}
We analyze \textit{GridAR} from three perspectives: (i) the trade-off between computational cost and performance, (ii) comparative performance of different verifier architectures, and (iii) the timing of prompt reformulation. Experiments are conducted under eight dimensions in T2I-CompBench++. We also explore failure cases of \textit{GridAR} in Appendix \textcolor{red}{E}.
\vspace{-0.11in}
\paragraph{Cost-performance analysis} 
While \textit{GridAR} shows notable improvements over Best-of-\textit{N}, it incurs additional computation due to the verification step - though this is substantially reduced by verifying four candidates at once. To better understand the trade-off between performance gains and computational overhead, we conduct a Pareto analysis, as shown in the Figure~\ref{fig:in_depth_anlysis} (left). Using Janus-Pro with two verifiers, GPT4.1 and MiniCPM-V 4.5~\citep{yu2025minicpm}, \textit{GridAR} (\textit{N}=4) achieves a 14.4\% and 11.7\% performance improvements, respectively, while reducing wall-clock time by 25.6\% and 27.6\% compared to Best-of-\textit{N} with \textit{N}=8. Specifically, the single-batch processing time (measured on RTX 3090 GPUs, including API latency) is 36.5s for GPT-4.1 and 35.5s for the MiniCPM-V 4.5 verifier, whereas Best-of-\textit{N} takes 23.9s (\textit{N}=4) and 49.1s (\textit{N}=8). Note that we optimize the wall-clock time of Best-of-N using such as KV-caching for a faithful comparison. These results demonstrate that \textit{GridAR} offers a more favorable cost-performance trade-off. A detailed breakdown of wall-clock time is in Appendix \textcolor{red}{D}.

\vspace{-0.12in}
\paragraph{Comparative evaluation of verifiers} 
As our framework exploits a verifier to assess partial image views and infer layouts in a zero-shot manner, it assumes a certain degree of image understanding capability from the verifier. In our experiments, we use GPT-4.1 as the default verifier; however, we also examine the performance of \textit{GridAR} (\textit{N}=4) with different verifier choices. As shown in the centered plot of Figure~\ref{fig:in_depth_anlysis}, GPT-4o-mini yields an 18.6\% improvement, and the open-source alternatives GLM-4V~\citep{glm2024chatglm} and MiniCPM-V~4.5~\citep{yu2025minicpm} deliver comparable gains of 17.6\% and 15.9\%, respectively, on text-to-image generation.
These results imply that our framework can further benefit from stronger vision-language models with enhanced image understanding capabilities. As this area continues to advance, we expect the upper bounds of our framework to rise with the emergence of more capable visual reasoning models.
\vspace{-0.11in}
\paragraph{Effect of prompt reformulation timing} 
A natural research question is whether reformulating the prompt \textit{before} image generation using a planner model could offer advantages over our strategy. While plausible, our approach instead performs prompt reformulation \textit{midway}, using layouts inferred from partial images. To compare, we generate images directly from our reformulated prompt (Figure~\ref{fig:in_depth_anlysis}, right).  While \textit{GridAR} ($N$=4) with initial prompt reformulation achieves an average score of 0.5018 in text-to-image generation - showing a clear improvement over no reformulation (average score 0.4753) - our layout-aware reformulation yields a substantially higher score of 0.5643. 
We attribute this to initial layout-guided prompts being susceptible to divergence from the intended layout, as early-stage next-token predictions may deviate from the structure. 

\vspace{-0.04in}
\section{Conclusion}
We have introduced \textit{GridAR}, a test-time scaling framework that rethinks how computation should be allocated in visual autoregressive (AR) models. Through progressive, grid-based generation and dynamic prompt reformulation, our approach selectively amplifies promising candidates while pruning suboptimal ones early - addressing limitations of conventional Best-of-\textit{N} strategies. Without requiring training, \textit{GridAR} draws out the full potential of AR models, achieving higher quality in both text-to-image generation and image editing. We believe this work marks a milestone for generative AR models, pushing the boundary of what test-time scaling can achieve in AR image generation.  

{
    \small
    \bibliographystyle{ieeenat_fullname}
    \bibliography{main}
}


\end{document}